\documentclass{article}

\usepackage[final,nonatbib]{neurips_data_2024}

\def\DRAFT

\usepackage[utf8]{inputenc} %
\usepackage[T1]{fontenc}    %
\usepackage{hyperref}       %
\usepackage{url}            %
\usepackage{booktabs}       %
\usepackage{amsfonts}       %
\usepackage{nicefrac}       %
\usepackage{microtype}      %
\usepackage{xcolor}         %

\usepackage{cite}
\usepackage{multirow}
\usepackage[export]{adjustbox}
\usepackage{pifont}
\newcommand{\tick}{\ding{52}}
\newcommand{\cross}{\ding{54}}
\usepackage[flushleft]{threeparttable} %
\usepackage{caption}

\title{WildPPG: A Real-World PPG Dataset of Long Continuous Recordings\\ }

\author{%
  Manuel Meier, Berken Utku Demirel, and Christian Holz%
    \\[.3em]
  Department of Computer Science\\
  ETH Z\"urich, Switzerland \\[.3em]
  \texttt{firstname.lastname@inf.ethz.ch} \\
}
\usepackage{bm}
\usepackage{mathtools}
\usepackage{amsmath}
\usepackage{multirow}
\usepackage{tabularx}
\usepackage{balance}
\usepackage{enumitem}
\usepackage{booktabs}
\usepackage[english]{babel}
\usepackage[autostyle,english=american]{csquotes}
\usepackage{graphics} %
\usepackage{threeparttable}
\usepackage{tikz}
\usetikzlibrary{automata,arrows,calc,positioning}
\usepackage{xspace}
\def\projname{WildPPG\xspace}

\ifx\DRAFT\fi

\begin{document}

\maketitle

\begin{abstract}
Reflective photoplethysmography (PPG) has become the default sensing technique in wearable devices to monitor cardiac activity via a person's heart rate (HR). 
However, PPG-based HR estimates can be substantially impacted by factors such as the wearer's activities, sensor placement and resulting motion artifacts, as well as environmental characteristics such as temperature and ambient light.
These and other factors can significantly impact and decrease HR prediction reliability.

In this paper, we show that state-of-the-art HR estimation methods struggle when processing \emph{representative} data from everyday activities in outdoor environments, likely because they rely on existing datasets that captured controlled conditions.
We introduce a novel multimodal dataset and benchmark results for continuous PPG recordings during outdoor activities from 16 participants over 13.5 hours, captured from four wearable sensors, each worn at a different location on the body, totaling 216\,hours.
Our recordings include accelerometer, temperature, and altitude data, as well as a synchronized Lead I-based electrocardiogram for ground-truth HR references.
Participants completed a round trip from Zurich to Jungfraujoch, a tall mountain in Switzerland over the course of one day.
The trip included outdoor and indoor activities such as walking, hiking, stair climbing, eating, drinking, and resting at various temperatures and altitudes (up to 3,571\,m above sea level) as well as using cars, trains, cable cars, and lifts for transport---all of which impacted participants' physiological dynamics.
We also present a novel method that estimates HR values more robustly in such real-world scenarios than existing baselines.

Dataset \& code for HR estimation: 
\textbf{\color{magenta}{\url{https://siplab.org/projects/WildPPG}}}

\end{abstract}

\section{Introduction}

Today's wearable devices such as smartwatches and fitness trackers commonly monitor a person's cardiac health by continuously assessing their heart rate (HR).
For estimating this metric, wearable devices predominantly use reflective photoplethysmography (PPG), which has become ubiquitous during ambulatory assessments due to its non-invasive nature and ease of use~\cite{nature_wearable}.
The HR measurements devices obtained from real-life conditions can supplement health assessments including exercise intensity, stress, fatigue, or sleep quality~\cite{JMIR, wearables_cvd, sleep_hr}.

However, obtaining \emph{reliable HR estimations} from PPG signals in real-world conditions and during real-world activities is challenging.
Several external factors negatively impact accurate estimation, such as motion artifacts~\cite{TROIKA} that arise from a person's movements and performed activities~\cite{HRV_UCI, Dalia}, sensor misplacement~\cite{lee_effective_2016} or slippage during wear, and environmental conditions that change over time such as low temperatures~\cite{khan_investigating_2015} or high levels of ambient light~\cite{kim_ambient_2015}.
Basic implementations of HR detection have been validated on comparably clean reference datasets, which do not adequately represent the variability and noise introduced by everyday activities and conditions.
Even for simple activities such as walking and running~\cite{poor_fitbit}, recent studies have provided evidence for the resulting detrimental effects on signal quality and, thus, the performance of HR tracking implementations.

To combat the effects of motions and activity on HR detection, prior work has proposed data-driven methods that learn to recover meaningful estimates under noisy conditions, including supervised methods~\cite{cardiogan, meier2024bsn, JBHI_ANC,Only_LSTM}, and rule-based techniques~\cite{TROIKA,IEEE_SPC2, meier2024embc, Temko, spa_original}.
While these and other methods estimate HR with high accuracy, their robustness to real-world effects is limited by the diversity of the datasets they were trained on.
Because existing datasets were dominantly captured in controlled conditions, they may not adequately represent the noisy nature under which today's wearables obtain PPG measurements in daily life and ``in the wild.''

\begin{figure}
    \centering
    \includegraphics[width=\columnwidth]{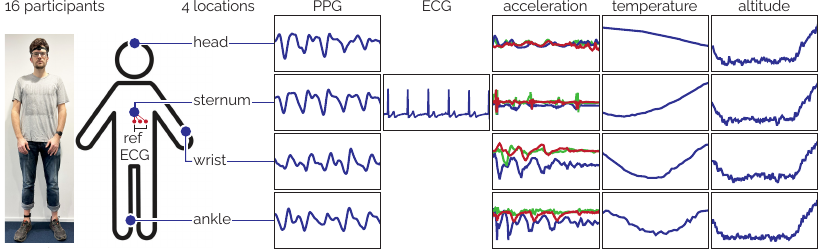}
    \caption{\projname comprises multi-modal signals from wearable devices at four sites on the body.
    Each device continuously recorded synchronized signals from a 3-channel reflective photoplethysmogram (red, green, infrared PPG), 3-axis inertial sensor (accelerometer), temperature, and barometric altitude sensor.
    For reference, the sternum device continuously recorded a Lead-I electrocardiogram (ECG) from body-mounted gel electrodes to provide ground-truth heart rate (HR) estimates.
    }
    \label{fig:data}
\end{figure}

In this paper, we introduce \projname, a novel dataset of reflective PPG measurements from four locations across the wearer's body and reference recordings from an electrode-based Lead-I electrocardiogram (ECG) for HR estimation tasks from everyday activity outside controlled environments.
Our dataset comprises continuous and synchronized recordings from 16 participants with an average duration of 13.5\,hours during a round-trip to a tall mountain in Europe, including various daytime activities and modes of transport.
As shown in Figure~\ref{fig:data}, our dataset additionally includes the continuous measurements from a 3-axis inertial sensor (accelerometer), barometric altitude sensor, and temperature sensor.
Our data collection simultaneously recorded all these signals from four devices across the wearer's body (wrist, ankle, sternum, head), totaling 216\,hours of synchronized multi-modal recordings that can enable a wide variety of future analyses.

We also contribute the implementation of multiple baseline methods for HR estimation and evaluate them on our dataset to identify weaknesses.
Based on our analysis, we introduce a learning-based method that takes temperature as input in addition to the raw PPG signal and show that its temperature awareness allows it to produce more robust HR estimates.

Collectively, we make the following contributions in this dataset paper:

\begin{itemize}[leftmargin=.5cm,nosep]
    \item a multi-modal real-world dataset of continuous, noisy, and synchronized PPG signals (raw red, green, infrared) from 4 body locations (forehead, sternum, wrist, ankle) and ECG recordings for reference,
    continuously collected from 16 participants over an average duration of 13.5\,hours (minimum duration: 12.3\,hours),

    \item a data collection study that covered everyday outdoor activities outside controlled (laboratory) conditions: During collection, participants completed a round-trip to a mountain station at over 3500\,meters above sea level using various modes of transportation (car, train, cable car, elevator) and engaging in various outdoor and indoor activities (walking, hiking, stair climbing, eating, drinking, social interactions, restroom breaks, and resting in varying conditions), including environmental conditions of low and medium temperatures and various degrees of solar radiation,

    \item an accuracy analysis of baseline methods that estimate HR on \projname based on environmental characteristics.
    We also contribute a learning-based method for HR estimation that takes a raw PPG signal and temperature as input and show that it produces more robust estimates, and

    \item additionally recorded and synchronized modalities in our dataset as the basis for future algorithm developments, including 3-axis inertial measurements (accelerometer), barometric altitude, and temperature data from each of the 4 body locations (some exposed, some covered by clothing).
\end{itemize}

\begin{table*}[b]
  \centering
  \caption{Qualitative comparison of \projname with previous datasets for HR estimation.}
  \label{tab:lit}
  \begin{threeparttable}
    \begin{adjustbox}{width=1\columnwidth,center}
    \renewcommand{\arraystretch}{1.2}
    \begin{tabular}{@{}llrccccc@{}}
    \toprule
    \textbf{\begin{tabular}[l]{@{}l@{}}Dataset\\ (year)\end{tabular}} & \textbf{\begin{tabular}[c]{@{}c@{}} Collection \\ methodology\end{tabular}} & \textbf{\begin{tabular}[c]{@{}c@{}}Hours\\ (total)\end{tabular}} & \textbf{\begin{tabular}[c]{@{}c@{}} Body-worn \\ locations \tnote{*}\end{tabular}} & \textbf{\begin{tabular}[c]{@{}c@{}} Wavelengths \\ (PPG) \end{tabular}} & \textbf{\begin{tabular}[c]{@{}c@{}} Multi- \\ modal\end{tabular}} & \textbf{\begin{tabular}[c]{@{}c@{}}In-the-wild\tnote{**}\\ dataset\end{tabular}} & \textbf{\begin{tabular}[c]{@{}c@{}}Altitude/Temperature\\ changes\end{tabular}} \\ \midrule
    \multirow{2}{*}{\textbf{Ours (2024)}} & \multirow{2}{*}{In the wild} & \multirow{2}{*}{216} & \{Head, Chest, & \multirow{2}{*}{\{Red, Green, Infra\}}  & \multirow{2}{*}{\tick}  & \multirow{2}{*}{\tick}  & \multirow{2}{*}{\tick}  \\
    & & & Wrist, Ankle\} & & & & \\ 
    Ear-PPG~\cite{montanari_earset} (2023) & Very controlled  & 17 & \{Ear\} & \{Red, Green, Infra\}  & \tick  & \cross  & \cross  \\
    Welltory~\cite{Welltory} (2021) & Very controlled  & 1 &  \{Wrist\}  & \{Red, Green, Blue\}  & \cross  & \cross  & \cross  \\
    DaLiA~\cite{Dalia} (2019) & Controlled  & 36 &  \{Wrist\}  & \{Green\}   & \tick  & \tick  & \cross  \\
    BAMI~\cite{BAMIDS} (2019) & Very controlled  & 10 &  \{Wrist\}  & \{Red, Green, Infra\}   & \cross  & \cross  & \cross  \\
    WESAD~\cite{WESAD} (2018) & Controlled  & 25 &  \{Finger\}  & \{Green\}  & \tick  & \cross  & \cross  \\
    BIDMC~\cite{bidmc} (2017) & Very controlled  & 7 &  \{Wrist\}  & ---  & \cross  & \cross  & \cross  \\
    PPGMotion~\cite{PPGMotion} (2017) & Very controlled  & 1.5 &  \{Wrist\}  & \{Green\}  & \cross  & \cross  & \cross  \\
    IEEE SPC~\cite{IEEE_SPC2} (2015) & Very controlled  & 2 &  \{Wrist\}  & \{Red, Green\}  & \cross  & \cross  & \cross  \\
    \bottomrule
    \end{tabular}
      \end{adjustbox}
     \begin{tablenotes}
  \item [*] \footnotesize \textit{Body-worn locations} refer to the sites on the body where PPG signals are measured. 
  \item [**] \footnotesize \textit{In-the-wild recording} is defined if the data were collected outside of (controlled) lab environments.
  \end{tablenotes}
  \end{threeparttable}
\end{table*}

\section{Related Work}

\subsection{PPG Datasets}
Monitoring the cardiac activity of people using blood volume pulse signals, i.e., photoplethysmography, has a long history in continuous mobile health monitoring~\cite{m_health}.
Robust HR estimation in real-world situations remains challenging due to the noisy nature of the PPG signals~\cite{TROIKA}.
In order to develop, evaluate, and compare novel approaches, publicly available datasets are essential. 
However, existing public datasets are limited in terms of data size and applicability to wearable devices in real-world scenarios. 
For example the IEEE Signal Processing Challenge (SPC) 2015 dataset~\cite{IEEE_SPC2}, includes 5-minute PPG recordings from 12 participants while the participants were controlled in the lab environments. 
This dataset is divided into two sets: a set with less motion artifacts from 12 participants (IEEE SPC12) and a set with heavy motions from 10 participants (IEEE SPC10).
The BAMI dataset~\cite{BAMIDS} was collected from 24 participants and includes resting, walking, and running activities.
The experiments consisted of 2 minutes of walking as a warm-up, 3 minutes of running, 2 minutes of walking, 3 minutes of running, and 2 minutes of walking to cool down, all on a treadmill inside lab environments where even the speed was controlled (6.0 to 7.0\,km/h).
Similarly, the PPGMotion dataset contains walking, running, and cycling activities of 8 participants with an average recording duration of 17 minutes~\cite{PPGMotion}. 
Meanwhile, the WESAD dataset~\cite{IEEE_SPC2} consists of approximately 1.6 hours of recordings for each of its 15 participants while they were seated and/or standing in a lab environment.
It does not contain any physical activities but includes stress and amusement stimuli as well as meditation. 
All aforementioned datasets were recorded in controlled indoor conditions.

To record data that is more representative of real-world conditions, i.e., outside of lab environments, the DaLiA dataset~\cite{Dalia} includes not only indoor but also outdoor activities such as cycling, walking, and driving.
It contains recordings of 15 participants with a total of 26 hours of PPG data. 
Although this dataset is the largest among existing datasets, it is approximately eight times smaller than our presented dataset. 
Moreover, while the activities in DaLiA are controlled outside of lab environments, our experiments were conducted in completely free-living conditions, offering a more realistic representation of everyday activities.
In other words, our dataset was collected "in the wild" ---participants were not controlled and were free to move or engage in activities of their choice (within the framework of the experimental protocol) outside of the lab environment.
It includes 216 hours of data, making it the largest dataset of its kind while including signals from multiple body locations with multiple wavelengths.
Moreover, it is multi-modal. Covering the varying altitude-temperature changes, ensuring comprehensive and complete coverage of real-life conditions.
Table \ref{tab:lit} qualitatively compares \projname with the existing PPG datasets.

\subsection{HR estimation methods}
Numerous efforts have investigated HR estimation from wearable devices using PPG signals during everyday motion and activity~\cite{phase_voco_ppg, Temko,imwut_ch,carek2018naptics}, given its crucial role in mobile health monitoring~\cite{wearables_cvd}.
Most of these studies employed Fourier transformation to observe how the frequency changes over time for estimating HR value~\cite{TROIKA, Curtoss, phase_voco_ppg}.
While this approach works for less contaminated PPG signals, motion artifacts hinder the measurement of heart rate in spectral density. 
As a result, many approaches have been proposed to obtain the heart rate using the power spectral densities of PPG and accelerometer signals to differentiate the frequency of heartbeats from motion artifacts~\cite{DeepPPG, sch_original,spa_original,phase_voco_ppg, JBHI_ANC}. 
However, if the dominant frequencies in accelerometer signals overlap with the true heart rate, it becomes challenging to distinguish signals in the frequency domain~\cite{pollreisz_detection_2022}.
In our previous work, we have shown that the combination of PPG signals from multiple body locations~\cite{meier2024embc,meier2024ciss_ppglocation} and the use of multiple PPG signals recorded at different wavelengths~\cite{meier2024bsn,meier2024ciss_ppgwavelength} improves HR estimations, though motion artifacts often manifest across body locations and affect all PPG sensors.
Our previous method BeliefPPG thus explicitly models uncertainties associated with PPG and IMU readings as well as estimated HR distribution, incorporating the statistical distribution of HR changes to refine estimates in a temporal context~\cite{bieri2023beliefppg}.

To enhance estimations based solely on PPG signals, recent works have proposed deep learning-based approaches~\cite{PPGnet, Binary_CorNet}. 
However, these models lack information about the degree of motion artifacts (MAs), leading to significant errors in heart rate estimation when inputs deviate from the training data distributions~\cite{DeepPPG}. 
Even methods that combine input from a PPG sensor and an accelerometer struggle to learn the intricate relationship between motion artifacts and blood volume flow signals, resulting in unreliable heart rate predictions~\cite{Temko}.
Thus, to address the limitations of existing PPG-based heart rate estimation methods, we introduce \projname which includes temperature as an additional modality alongside PPG recordings, resulting in improved heart rate estimation accuracy.

\begin{figure}
    \centering
    \includegraphics[width=\columnwidth]{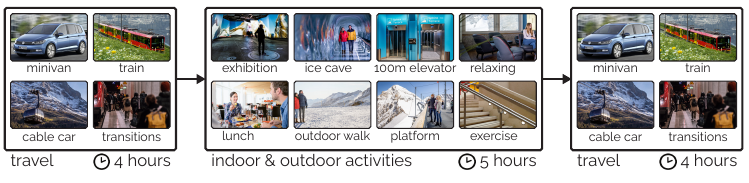}
    \caption{\projname\ participants engaged in multiple forms of travel as well as indoor and outdoor activities with changing environmental conditions.
    No strict study protocol was enforced and participants completed the activities at their own preferred speed.}
    \label{fig:procedure}
\end{figure}

\section{\projname Dataset}

The purpose of \projname is to enable the development of more reliable PPG-based algorithms that improve the robustness of estimating HR as well as potential other cardiovascular metrics that are of interest on wearable devices such as smartwatches or fitness trackers.
Our data recording methodology was designed to capture real-world data under less-than-optimal conditions during representative outdoor activities and everyday conditions.
To achieve this, \projname contains long and uninterrupted recordings that our apparatus recorded from participants outside controlled environments or procedures.
While our primary focus was on recording the signals crucial for HR estimation---PPG time series from a reflective green light-based sensing design across multiple body locations as well as chest ECG-based cardiac activity measurements for reference---our apparatus additionally recorded motion data from an inertial measurement unit at each body site, supplementary PPG signals from red and infrared reflections, as well as temperature readings and barometric altitude measurements throughout the recording for each participant.

Table~\ref{tab:lit} qualitatively compares our dataset's characteristics with commonly used datasets for HR prediction from blood volume pulse signals.
We specifically focus on comparing our dataset with those collected under conditions similar to real-world conditions, where participants moved freely.
BIDMC is an exception, as it was collected with critically ill adults (a subset of the MIMIC-II dataset), and we include it because of its use in learning-based HR prediction in previous work (e.g.,~\cite{cardiogan}).
As shown in Table~\ref{tab:lit}, our dataset is $\sim8\times$ bigger than existing datasets for HR prediction and includes multi-modal data collected under real-world conditions.
Below, we detail the experimental protocol and the design of our data capture apparatus for \projname.

\subsection{Dataset Design}
\subsubsection{Experimental Protocol}
Participants gathered in the morning to start the study.
An experimenter outfitted each participant with four recording devices and ensured PPG and ECG signal quality (20\,min).
The participants then took a minivan from Zurich to Grindelwald (140\,min) and transitioned to a cable car and train to Jungfraujoch railway station at 3460\,m above sea level (80\,min).
Subsequently, they spent 5\,hours in public places in and around the station and engaged in various activities.
While the stay at the station included the same main activities for all participants, no strict study protocol was enforced in an effort to capture real-world data while participants completed the activities at their own speed of preference.
Participants were encouraged to not constrain themselves due to the study protocol and engage in social interactions, take pictures, buy and consume snacks and beverages, take restroom breaks as needed, or sit down throughout the stay.

As shown in Figure~\ref{fig:procedure}, the main activities included walking through the museum and exhibition area including an ice cave with below-freezing temperatures ($\approx$ 60\,min), taking a 110-meter-high elevator and walking the remaining stairs up to the observatory and outside viewing platform ($\approx$ 60\,min), sitting down for lunch ($\approx$ 60\,min), walking through the snow-covered outside area ($\approx$ 60\,min), and resting inside ($\approx$ 60\,min). 
Spread out throughout the stay at Jungfraujoch, the participants additionally climbed and descended a set of stairs across four floors at maximum endurable pace on three to four occasions.
After the stay at the station, participants took the train and cable car back to Grindelwald (80\,min) and returned to Zurich on the minivan (140\,min).
Upon arrival, the experimenter removed the recording devices from each participant.
Throughout the study, participants were instructed to remove the device on the wrist when washing their hands.
All other devices were worn continuously.

\subsubsection{Sensors}
\begin{figure}
    \centering
    \includegraphics[width=\columnwidth]{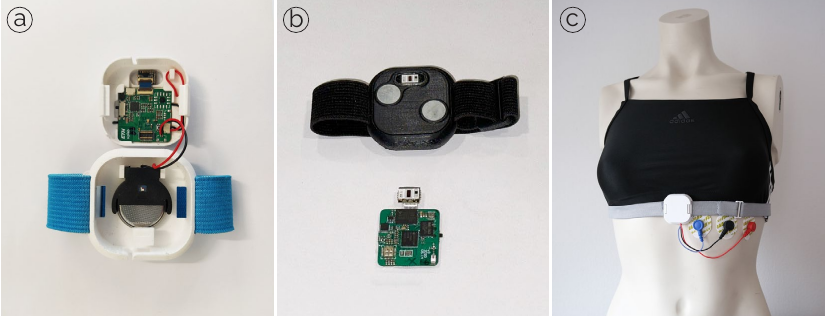}
    \caption{The wearable devices used for the data recording were custom-built and are centered around a SoC to read and store all synchronized sensor data in on-board flash memory.
    (a)~Devices were powered by a CR2032 coin cell batteries (runtime: 18\,hours) inside a 3D-printed case with (b)~a flexible and adjustable strap.
    (c)~The sternum unit additionally connected to 3 gel-electrodes to obtain a continuous ECG recording for reference.}
    \label{fig:device}
\end{figure}

The signals collected in \projname were acquired using miniaturized, custom-built low-power wearable devices.
As shown in Fig.\,~\ref{fig:device}, the devices are encased in a 3D-printed body and have an adjustable, stretchable strap with which they were attached at the forehead, sternum, ankle (supramalleolar), and wrist (dorsal).
PPG measurements were obtained using an optical analog front-end at 128\,Hz (MAX86141, Analog Devices) that connected to an optical module (SFH7072, ams-OSRAM) with a green (530\,nm), a red (660\,nm), and an infrared (950\,nm) LED as well as a broadband photodiode (410 -- 1100\,nm), and an infrared-cut photodiode (402 -- 694\,nm).
Green and red PPG were acquired in combination with the infrared-cut photodiode and infrared PPG with the broadband photodiode.
Accelerometer data was acquired at a sampling rate of 200\,Hz using a MEMS digital motion sensor (LIS2DH, STMicroelectronics).
For ground truth, the sternum device additionally collected the Lead~I ECG at 128\,Hz through a biopotential sensor (MAX30003, Analog Devices) that connected to gel electrodes placed on the chest.
Temperature and barometric altitude measurements were collected at a sampling rate of 10\,Hz within the device (BME280, Bosch Sensortec).

Raw sensor data was continuously read from the sensors' FIFOs in batches by a System-on-a-Chip (DA14695, Dialog Semi), timestamped, stored in NAND memory (TH58CYG3S0HRAIJ, Kioxia Corp.), and downloaded after the recording was completed.
Checksums were used to guarantee correct data download.
Each device was powered by a CR2032 coin cell battery.

\subsubsection{Synchronization}
Since proper synchronization between sensors and devices is critical for this dataset, we collected the measurements \enquote{as synchronized as possible} on an electronics level and ensured proper trace alignment beyond the timing tolerances of the sensors.
The most important measurements for this dataset, PPG and ECG, were triggered by an external clock derived from the real-time clock (RTC) of the System-on-a-Chip and thus perfectly synchronized with the system time of the device.
This reduced the sample rate tolerance from $\pm 2\%$ of the internal sensor clock~\cite{max86141} by a factor of 100 to the tolerance of the used quartz (ECS-.327-7-16-C-TR, ECS Inc., $\pm 20$ppm tolerance~\cite{ecs_quartz}).
Temperature and barometric pressure measurements were triggered based on the same RTC and therefore also perfectly synchronized to the PPG and ECG measurements.
The accelerometer sampled based on its own internal clock.
Deviations in sampling rate due to clock tolerances were corrected in post-processing based on the RTC timestamps of the FIFO readouts of the sensor.

This setup guarantees synchronization for the different sensing modalities acquired by a single device.
However, measurements across devices are still participant to clock skew and drift effects -- most notably when comparing PPG measurements from the wrist, ankle, and head devices with ground truth ECG measurements acquired on the chest.
With the given quartz tolerance, although low, two devices may experience sampling shifts of up to 2\,seconds across a 14-hour continuous recording session.
Therefore, sensor data across devices was synchronized by aligning recorded signals offline following the approach described by Meier and Holz (33\,ms accuracy)~\cite{meier_bmar_2023}.

\subsection{Recruitment and Recording}
\label{subsec:recruitment}
\projname's participants were 16 healthy adults, recruited on a voluntary basis without compensation beyond the coverage of all expenses incurred throughout the recording of the data and the visit of the mountain station.
7 female and 9 male participants took part, with ages ranging between 22 and 69 (mean age: 39).
All participants' skin tones were within 1--3 on the Fitzpatrick scale~\cite{fitzpatrick1988validity}.
All participants signed a consent form prior to participation.
The experimenters conducted the data collection with groups of 3--5\,participants, and the protocol was identical for all participant groups.

At the beginning of the recording, participants were introduced to the experimental protocol and the goal of the study.
An experimenter was always with the participants throughout the recording.
The recording devices worked without requiring user input, operating in a completely passive manner.
The apparatus provided no feedback of any sort to participants either.

\subsubsection{Risks}
\label{subsubsec:risks}
Participants were informed about the risk of developing symptoms of Acute Mountain Sickness (AMS) during the visit to Jungfraujoch station due to its altitude.
Trains leaving to lower altitudes were available every 30\,minutes in case a participant was feeling unwell.
Additionally, a professional first aid station is available at Jungfraujoch alongside trained personnel and medical assistance if needed.
The study protocol and risk assessment were approved by the ethics committee of ETH Zürich (EK 2022-N-44).

\begin{table*}[t]
    \centering
    \caption{The synchronized time series captured in \projname.}
    \begin{tabular}{@{} lllr  cccc @{}}
        \multirow{2}{*}{\textbf{Time Series}} & \multirow{2}{*}{\textbf{Description}} & \textbf{depth} & \multicolumn{1}{c}{\textbf{fps}} & \multirow{2}{*}{\textbf{head}} & \multirow{2}{*}{\textbf{sternum}} & \multirow{2}{*}{\textbf{wrist}} & \multirow{2}{*}{\textbf{ankle}} \\
         & & [bits] & [Hz]  &  &  & & \\
        \midrule
        \texttt{ppg\_g}      & Green PPG     & 19 & 128 & $\times$ & $\times$ & $\times$ & $\times$ \\
        \texttt{ppg\_ir}     & Infrared PPG  & 19 & 128 & $\times$ & $\times$ & $\times$ & $\times$ \\
        \texttt{ppg\_r}      & Red PPG       & 19 & 128 & $\times$ & $\times$ & $\times$ & $\times$ \\
        \texttt{ecg}         & Lead I ECG  & 18 & 128 & & $\times$ \\[.5ex]
        \texttt{accel\_x}    & Accelerometer X-axis   & 10 & 128 & $\times$ & $\times$ & $\times$ & $\times$ \\
        \texttt{accel\_y}    & Accelerometer Y-axis   & 10 & 128 & $\times$ & $\times$ & $\times$ & $\times$ \\
        \texttt{accel\_z}    & Accelerometer Z-axis   & 10 & 128 & $\times$ & $\times$ & $\times$ & $\times$ \\
        \texttt{altitude}    & Barometric altitude  & 23 & 0.5 & $\times$ & $\times$ & $\times$ & $\times$ \\
        \texttt{temp}        & Temperature (inside case)  & 16 & 0.5 & $\times$ & $\times$ & $\times$ & $\times$ \\
        \bottomrule
    \end{tabular}
    \label{tab:sensors}
\end{table*}

\subsection{Dataset Composition}

Participants' recordings are limited to the time they actually wore the sensor devices.
PPG and ECG recordings are available as continuous recordings of raw sensor data.
Accelerometer recordings were downsampled and synchronized to match the 128\,Hz of the PPG and ECG signals.
Device temperature measurements were averaged with an 8-second sliding window (2-second step size).
The conversion from barometric pressure measurements to altitude above sea level was calculated following the conversion formula in the sensor's datasheet and a normed barometric pressure measurement from the same day acquired by a weather station at Jungfraujoch station run by the Swiss Federal Office of Meteorology and Climatology~\cite{meteoswiss}.
Barometric altitude measurements were averaged across all 4\,devices worn by a participant and averaged with an 8-second sliding window and 2-second step size.
Overall, \projname contains 216\,hours of synchronized recordings.
Accounting for the four separate device locations, this corresponds to 864\,hours of PPG recordings at 3 different wavelengths.
An overview of all captured data is shown in Table~\ref{tab:sensors}.

\subsubsection{Ground Truth}
\label{subsec:gt}
As manual annotation of a dataset of this size is impractical and unreliable, the dataset contains Lead I-based ECG recordings for ground truth reference.
For baseline analysis, we detected R-peaks in the ECG signal, using Pan-Tompkins~\cite{pan_real-time_1985}.
To reduce the risk of corrupt ground truth values, peaks with inter-beat intervals (IBI) corresponding to HR values greater than 185\,bpm or less than 35\,bpm were removed, and for HR calculation, only IBIs that were part of a sequence of 4 IBIs for which $IBI_{min}/IBI_{max}<0.75$ were considered.
In data windows with less than two remaining IBI, no ground truth HR is computed and the window is omitted from baseline methods.
Across the whole dataset, this applies to 2640 8-second windows (2.7\% of total) mainly due to three participants with partially noisy ECG recordings which are responsible for 2164 of the rejected windows.

\section{Baselines}
We computed 6 heuristic and 5 supervised baseline algorithms as well as our own method on \projname. 
To improve comparability, all baselines were computed on data recorded by wrist-worn devices.
All results are shown in Table~\ref{tab:performance_ppg} and the methods are described in more detail in the following two subsections.
\subsection{Heuristic Methods}
\label{subsec:heuristicmethods}
MSPTD is a peak detection method suited for PPG signals that computes a local maxima scalogram (LMS), a matrix that contains information about local maxima computed at different scales by comparing samples to neighbors at varying distances in the signal. 
The concept was first published by Scholkmann et al~\cite{scholkmann_efficient_2012} and expanded upon by Bishop and Ercole by combining maxima detection with minima detection and improvements in algorithm efficiency~\cite{bishop_multi-scale_2018}. The algorithm does not require any parameter tuning and works for any periodic or quasi-periodic signal.
qppg is another peak detection algorithm built specifically for PPG signals and works by integrating the positive slope between diastolic and systolic points in the signal. Peaks are detected using adaptive thresholding \cite{vest_open_2018}.
HeartPy detects peaks in the PPG signal where it crosses its own moving average plus a variable offset. Multiple such offsets are dynamically tested and the one chosen which produces the most regularly spaced detected beats \cite{van_gent_heartpy_2019}.
PWD is another PPG-specific method that relies on the detection of beats based on zero-crossings in the first derivative of the signal~\cite{li_automatic_2010}.
MSPTD, HeartPy, qppg, and PWD methods were computed using the MATLAB implementation by Charlton et al. which is released under GPL-3.0 open-source license~\cite{charlton_detecting_2022}.
We also included traditional Fourier transformation and autocorrelation functions, where both are used to detect periodicity~\cite{saul_periodic}.

\begin{table*}[b]
\centering
\caption{Performance comparison of baselines with prior works in datasets}
  \begin{threeparttable}
\begin{adjustbox}{width=1\columnwidth,center}
\renewcommand{\arraystretch}{1.1}
\begin{tabular}{@{}lllllllllll@{}}
\toprule
\multirow{2}{*}{Method} & \multicolumn{3}{l}{WildPPG} & \multicolumn{3}{l}{SPC12\tnote{*}} & \multicolumn{3}{l}{DaLiA} \\ 
\cmidrule(r{15pt}){2-4}  \cmidrule(r{15pt}){5-7}  \cmidrule(r{15pt}){8-10}
& MAE$\downarrow$ & RMSE$\downarrow$ & $\rho$$\uparrow$ & MAE$\downarrow$ & RMSE$\downarrow$ & $\rho$$\uparrow$ & MAE$\downarrow$ & RMSE$\downarrow$ & $\rho$$\uparrow$ \\
\midrule
\textit{Heuristic} & & & & & & & & & \\
FFT & 17.62 & 28.68 & 12.11 & 14.83 & 25.81 & 35.15 & 34.98 & 47.13 & 2.72\\
Autocorrelation & 28.43 & 35.23 & -3.06 & 41.16 & 49.81  & 3.945 & 30.59 & 39.09 & 9.18 & \\
HeartPy & 18.49 & 30.15 & 14.7 & 19.89 & 24.81 & 20.19 & 12.25 & 18.15 & 74.97 \\
MSPTD & 13.30 & 21.19 & 21.37 & 20.35 & 25.31 & 18.22 & 17.20 & 25.11 & 53.95 \\
PWD & 11.72 & 19.57 & 25.19 & 21.16 & 26.32 & 18.75 & 28.26 & 37.48 & 29.51 \\
qppgfast & 19.32 & 29.94 & 21.27 & 17.97 & 21.99 & 32.72 & 13.31 & 20.35 & 66.58 \\
\midrule
\textit{Self-supervised} & & & & & & & & & \\
SimCLR & 15.75\small$\pm$1.81 & 19.81\small$\pm$1.17 & 8.12\small$\pm$2.51   & 12.42\small$\pm$0.05 & 20.96\small$\pm$0.30 & 60.41\small$\pm$0.52                & 12.01\small$\pm$0.14 & 19.46\small$\pm$0.14 &58.31\small$\pm$0.39 & \\
NNCLR  & 14.46\small$\pm$0.13 & 18.43\small$\pm$0.05 & 11.54\small$\pm$0.56             & 13.14\small$\pm$0.49 & 18.86\small$\pm$0.49 & 69.82\small$\pm$0.06                & 12.94\small$\pm$0.31 & 20.02\small$\pm$0.49 & 51.12\small$\pm$2.54 \\
BYOL   & 14.11\small$\pm$0.12 & 18.42\small$\pm$0.15 & 12.50\small$\pm$0.50             & 18.71\small$\pm$0.93 & 25.01\small$\pm$1.50 & 48.82\small$\pm$4.36                & 11.67\small$\pm$0.32 & 17.57\small$\pm$0.23 &63.96\small$\pm$0.97 & \\
TS-TCC & 12.64\small$\pm$0.03 & 17.72\small$\pm$0.04 & 18.71\small$\pm$0.41             & 11.56\small$\pm$0.41 & 18.04\small$\pm$0.66 & 68.38\small$\pm$1.41                & 8.12\small$\pm$0.30 & 14.89\small$\pm$0.21 & 67.13\small$\pm$0.53 \\
TS2Vec & 10.55\small$\pm$0.37  & 16.52\small$\pm$0.39 & 26.31\small$\pm$0.25            & 9.75\small$\pm$0.08  & 17.82\small$\pm$0.43 & 75.43\small$\pm$0.33                & 10.83\small$\pm$0.13 & 17.89\small$\pm$0.19 & 60.10\small$\pm$0.62 \\
VICReg & 15.37\small$\pm$0.60 & 19.20\small$\pm$0.33 & 9.30\small$\pm$1.57             & 13.17\small$\pm$0.82 & 20.38\small$\pm$1.27 & 59.76\small$\pm$4.16                & 14.90\small$\pm$0.16 & 21.94\small$\pm$0.11 & 45.38\small$\pm$0.07\\              
Barlow Twins & 16.14\small$\pm$0.92 & 20.21\small$\pm$0.90 & 9.13\small$\pm$1.20 & 13.22\small$\pm$0.34 & 20.42\small$\pm$0.88 & 64.51\small$\pm$4.01 & 18.26\small$\pm$0.57 & 23.41\small$\pm$0.29 & 23.42\small$\pm$7.30\\
\midrule
\textit{Unsupervised} & & & & & & & & & \\
UPD & 23.12\small$\pm$1.10 & 25.10\small$\pm$0.56 & 7.39\small$\pm$1.83 & 9.30\small$\pm$0.10 & 16.50\small$\pm$0.20 & 77.60\small$\pm$0.43 & 27.41\small$\pm$4.73 & 31.26\small$\pm$4.55 & 18.12\small$\pm$3.86 \\
\bottomrule
\textit{Supervised} & & & & & & & & & \\
1D ResNet & \underline{8.62}\small$\pm$0.06 & \underline{15.00}\small$\pm$0.06 & \underline{42.41}\small$\pm$0.24 & \textbf{5.50}\small$\pm$0.29 & \textbf{11.16}\small$\pm$0.64 & \textbf{84.86}\small$\pm$0.51 & \underline{4.47}\small$\pm$0.03 & \underline{10.03}\small$\pm$0.10 & \underline{85.87}\small$\pm$0.25\\
DCL      & 8.64\small$\pm$0.01 & 14.73\small$\pm$0.05 & 41.39\small$\pm$0.11 & 16.50\small$\pm$1.42 & 21.61\small$\pm$1.56 & 19.90\small$\pm$1.10 & 26.34\small$\pm$1.10 & 25.53\small$\pm$4.2 & 80.57\small$\pm$0.09 \\
FCN      & 9.79\small$\pm$0.23 & 15.62\small$\pm$0.26 & 35.68\small$\pm$0.94 & 27.55\small$\pm$0.55 & 31.18\small$\pm$0.81 & 40.41\small$\pm$1.81 & 6.55\small$\pm$0.28 & 11.24\small$\pm$0.32 & 83.21\small$\pm$0.04 \\
LSTM & 9.28\small$\pm$0.22 & 15.14\small$\pm$0.11 & 36.96\small$\pm$1.21 & 18.66\small$\pm$3.49 & 25.02\small$\pm$3.45 & 56.23\small$\pm$4.84 & 5.30\small$\pm$0.05 & 11.10\small$\pm$0.15 & 78.99\small$\pm$0.44 \\
Transformer   & 10.06\small$\pm$0.16 & 16.23\small$\pm$0.20 & 32.07\small$\pm$1.60 & 22.00\small$\pm$0.23 & 27.30\small$\pm$0.29 & 52.86\small$\pm$0.10 & 7.84\small$\pm$0.11 & 15.06\small$\pm$0.17 & 68.62\small$\pm$0.53 \\
Temp-ResNet    & \textbf{8.37}\small$\pm$0.02 & \textbf{14.72}\small$\pm$0.01 & \textbf{45.50}\small$\pm$0.14 & --- & ---&---& \textbf{4.40}\small$\pm$0.02 & \textbf{9.87}\small$\pm$0.11 & \textbf{85.90}\small$\pm$0.17 \\
\bottomrule
\end{tabular}
\end{adjustbox}
\begin{tablenotes}
  \item [*] \footnotesize The dataset has no temperature value from the measurement site.
  \end{tablenotes}
\end{threeparttable}
\label{tab:performance_ppg}
\end{table*}

\subsection{Supervised Methods}
For each supervised baseline, we follow the original implementation of the architectures.
Additionally, we search for the best hyperparameters (learning rate and its scheduler, batch size, weight decay).
Specifically, we compared our method with 1D ResNet~\cite{resnet1d} which is a modified version of ResNet architecture~\cite{ResNet} for time series with 1D filters.
DCL~\cite{KDD_paper} is a widely used network that combines convolutional and long-short-term memory units for HR prediction from PPG signals~\cite{KDD_paper, CorNET}.
Fully convolutional neural networks (FCN) which only include convolutional blocks with ReLU non-linearities.
LSTM~\cite{LSTM} model which is designed to capture temporal dependencies in time series data through its memory cell architecture.
Transformer~\cite{attention_is_all_you_need} is a widely adopted deep learning model that employs self-attention mechanisms to capture long-range dependencies.

\begin{figure}
    \centering
    \includegraphics[width=\columnwidth]{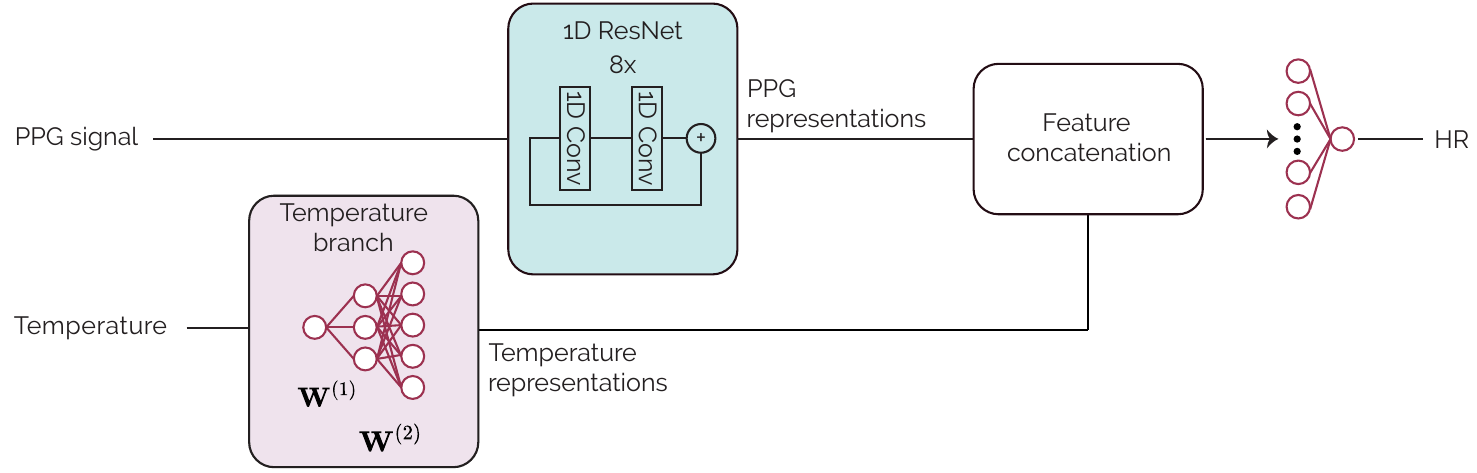}
    \caption{Architecture of Temp-ResNet. The ResNet includes batch normalization and ReLU activations after each convolution.}
    \label{fig:tempresnet}
\end{figure}

\subsubsection{Proposed Method: Temp-ResNet}
Considering the relation between the temperature of the measurement site and the SNR of PPG signals, we modified the 1D ResNet model by incorporating a small branch that inputs the temperature value of the segment to the overall architecture for HR estimation. 
Specifically, we use a multilayer perceptron with one hidden layer to obtain the representation for temperature \( \mathbf{z}_{temp} = \mathbf{W}^{(2)} \sigma(\mathbf{W}^{(1)} \mathbf{x}_{temp}) \) where \( \sigma \) and $\mathbf{x}_{temp}$ are a ReLU nonlinearity and temperature value of the specific segment, respectively.
The weight matrices \(\mathbf{W}^{(1)}\) and \(\mathbf{W}^{(2)}\) are learnable parameters of the MLP, responsible for mapping the input temperature value to a higher-dimensional space. Initially, the temperature value \(\mathbf{x}_{temp}\) is expanded to a 5-dimensional vector through the first layer, \(\mathbf{W}^{(1)} \in \mathbb{R}^{5 \times 1}\). 
This 5-dimensional representation is then further expanded to a 10-dimensional vector by the second layer, \(\mathbf{W}^{(2)} \in \mathbb{R}^{5 \times 10}\). 
Then, the obtained representation is concatenated with the extracted PPG representations and fed to the final linear layer for HR estimation. The architecture of Temp-ResNet is shown in Figure~\ref{fig:tempresnet}.

\section{Limitations}
\label{sec:limitations}
Regarding skin pigmentation, all participants were within 1--3 on the Fitzpatrick scale, and no participants rated themselves 4--6 (darker skin tones).
Considering the more challenging nature of PPG measurements among people with darker skin tones~\cite{fallow2013influence,bickler2005effects}, it will be important to broaden data collection in future efforts.
More generally, including 16 participants is a limitation of our dataset, and including more would allow covering as many participant-specific effects as possible.
Lastly, as described in Section~\ref{subsec:gt}, 2.7\% of the total dataset is affected by noisy ground truth measurements.

\section{Ethical Considerations}
\label{sec:ethics}
Working with real-world physiological and activity data from humans requires ethical considerations.
All participants were informed regarding safety and privacy both verbally and in writing.
They signed a consent form that they received prior to the study for careful reading.
No personal information is released with the dataset, the data is fully anonymous and does not contain personally identifiable information or offensive content.
To the best of our knowledge, no potential harm or malicious use is possible using the \projname dataset and the provided methods.
The study was approved by the ethics committee of ETH Zürich.

\section{Conclusion}
We have presented \projname, a real-world dataset of long and continuous multimodal motion and physiology recordings for PPG-based HR analysis.
Our experimental protocol included a wide variety of activities and environmental conditions including indoor and outdoor activities at different ambient temperatures, altitude levels, and light conditions.
Participants moved, acted, and interacted naturally throughout the day and recording without being bound to a strict study procedure---for the purpose of capturing representative data that resembles the measurements taken by wearable devices in \emph{everyday} use.
Besides PPG measurements at the three common red, green, and infrared wavelengths and ECG ground-truth references, \projname includes acceleration, temperature, and barometric altitude measurements to suit a wide range of analysis methods and tasks.
Our study recorded and synchronized measurements from multiple body locations to support exploring future estimation techniques that exceed the analysis of signals from wrist-worn devices most common on smartwatches today and to capture cardiac activity more holistically and robustly in the future.

To the best of our knowledge, \projname is the largest available dataset of its kind and can enable learning pipelines that benefit from longer time series during training, additional modalities, and real-world data (``in the wild'').
Beyond the dataset, \projname includes a range of baseline algorithms---heuristic as well as supervised---and our novel method that leverages temperature readings for improved HR estimation.
These methods can serve as a starting point for future work and benchmarks and our intention is to enable future work to better tailor methods to real-world situations and applications in less-than-optimal conditions.
This could include multi-modal processing or analyses of additional cardiovascular metrics that are of interest in wearable devices.

\begin{ack}
This work has not received any third-party funding and the authors declare no conflicts of interest.

\end{ack}

\bibliographystyle{IEEEtran}
\bibliography{refs}
\newpage
\section*{Checklist}

\begin{enumerate}

\item For all authors...
\begin{enumerate}
  \item Do the main claims made in the abstract and introduction accurately reflect the paper's contributions and scope?
    \answerYes{}
  \item Did you describe the limitations of your work?
    \answerYes{see Section~\ref{sec:limitations}}
  \item Did you discuss any potential negative societal impacts of your work?
    \answerYes{see Section~\ref{sec:ethics}}
  \item Have you read the ethics review guidelines and ensured that your paper conforms to them?
    \answerYes{}
\end{enumerate}

\item If you are including theoretical results...
\begin{enumerate}
  \item Did you state the full set of assumptions of all theoretical results?
    \answerNA{}
	\item Did you include complete proofs of all theoretical results?
    \answerNA{}
\end{enumerate}

\item If you ran experiments (e.g. for benchmarks)...
\begin{enumerate}
  \item Did you include the code, data, and instructions needed to reproduce the main experimental results (either in the supplemental material or as a URL)?
    \answerYes{}
  \item Did you specify all the training details (e.g., data splits, hyperparameters, how they were chosen)?
    \answerYes{Section~\ref{appendix_baselines}}
	\item Did you report error bars (e.g., with respect to the random seed after running experiments multiple times)?
    \answerYes{}
	\item Did you include the total amount of compute and the type of resources used (e.g., type of GPUs, internal cluster, or cloud provider)?
    \answerYes{Section~\ref{appendix_baselines}}
\end{enumerate}

\item If you are using existing assets (e.g., code, data, models) or curating/releasing new assets...
\begin{enumerate}
  \item If your work uses existing assets, did you cite the creators?
    \answerYes{\ref{subsec:heuristicmethods}}
  \item Did you mention the license of the assets?
    \answerYes{\ref{subsec:heuristicmethods}}
  \item Did you include any new assets either in the supplemental material or as a URL?
    \answerYes{}
  \item Did you discuss whether and how consent was obtained from people whose data you're using/curating?
    \answerYes{\ref{sec:ethics}}
  \item Did you discuss whether the data you are using/curating contains personally identifiable information or offensive content?
    \answerYes{\ref{sec:ethics}}
\end{enumerate}

\item If you used crowdsourcing or conducted research with human subjects...
\begin{enumerate}
  \item Did you include the full text of instructions given to participants and screenshots, if applicable?
    \answerNA{Not applicable.}
  \item Did you describe any potential participant risks, with links to Institutional Review Board (IRB) approvals, if applicable?
    \answerYes{\ref{subsubsec:risks}}
  \item Did you include the estimated hourly wage paid to participants and the total amount spent on participant compensation?
    \answerYes{\ref{subsec:recruitment}}
\end{enumerate}

\end{enumerate}

\newpage
\appendix

\section*{Appendix}
\section{Dataset Accessibility}
We provide \projname time series in the MATLAB .mat file format.
Code is provided in Matlab .m file format and Python .py source code. 
The python source code contains functions to load the .mat files into equivalent python structures.
All data, code, and the project website are hosted on ETH Zurich servers with long-running maintenance intended for long-term availability.
The dataset uses a Creative Commons CC BY-NC-SA 4.0 license~\cite{cc_by_nc_sa} while the code is released under GPL-3~\cite{gpl3}.

\section{Experimental Settings}
This section provides more details on the implementations of the learning-based baseline methods and the dataset we tested them on.
\subsection{Datasets}
\label{appendix_datasets}
The datasets used in this study are both widely employed in related work about HR estimation from PPG signals.

\paragraph{IEEE SPC}
This competition provided a training dataset of 12 participants (SPC12) and a test dataset of 10 participants~\cite{TROIKA}.
The IEEE SPC dataset overall has 22 recordings of 22 participants (ages 18 -- 58) performing three different activities. 
Each recording contains data sampled at 125\, Hz from a 3-axis accelerometer and 2-channel pulse oximeter sensor (green LEDs). 
All these recordings were captured from the wearable device placed on the wrist of each individual.
Additionally, a chest ECG provides ground-truth HR estimation. 
During our experiments, we used the PPG channels only. 
We use leave-one-out-cross-validation similar to the previous setups.

\paragraph{DaLiA}
PPG signals from the DaLia dataset were recorded at a sampling rate of 64 Hz. 
We used the whole dataset of 15 participants while following the leave-one-out-cross-validation.
10\% of the training data is randomly split for validation, i.e., early stopping and saving the best model.

All PPG datasets are standardized as follows.
Initially, a fourth-order Butterworth bandpass filter with a frequency range of 0.5–4 Hz is applied to the PPG signals. 
Subsequently, a sliding window of 8 seconds with 2-second shifts is employed for segmentation, followed by z-score normalization of each segment. Lastly, the signal is resampled to a frequency of 25 Hz for each segment.

\subsection{Supervised Baseline Methods}
\label{appendix_baselines}
We train the models using a batch size of 128 with a learning rate of $5e-4$.
The validation loss is employed to save the optimal model and halt training to prevent overfitting. 
If no improvements are observed for 15 consecutive epochs, we reduce the learning rate by a factor of 10.
Our experiments utilized NVIDIA GeForce RTX 4090 GPUs and involved training with three random seeds across all datasets, resulting in approximately 480 GPU hours, including ablation.

\paragraph{1D ResNet} is a modified version of ResNet for time series with 1D filters of size five across eight residual blocks.
The initial block consists of 32 filters, incrementing by a factor of two for every two subsequent residual blocks.
Also, batch normalization~\cite{batch} is applied after each convolutional block.
We apply a pooling operation after each residual with a stride of 2 while using Dropout~\cite{Dropout} with 0.5 after each activation.
Finally, a global average pooling is used before the final linear layer.

\paragraph{DCL} is a combination of convolutional blocks with the long-short term memory units~\cite{LSTM} (LSTMs) and is widely used for time series~\cite{KDD_paper, CorNET} as it considers the temporal relationship.
Specifically, the DCL architecture has four convolutional layers with $5\times1$ size of 64 kernels.
Then, the output is fed into the 2-layer LSTM with 128 units.
The last time step of the LSTM is fed to a linear layer.

\paragraph{FCN}
We also implemented a fully convolutional neural network with a 3-layer followed by ReLU activation and
MaxPooling after each convolutional layer, similar to the implementation in~\cite{KDD_paper}.
Dropout with 0.5 is applied after the first convolutional layer. 
We set the kernel and padding size to 8 and 4, respectively for each convolutional layer. 
The number of kernels for each convolutional layer is set to 32 for the first one and 64 for the rest.

\paragraph{LSTM}
We employed a unit of two layers of LSTM~\cite{LSTM} without any feature extraction, i.e., the raw PPG data is fed to the model. 
The features from the final LSTM are fed to a linear layer for HR estimation.
\paragraph{Transformer}
We also added transformers with positional encodings as a baseline model.
Since the attention mechanism~\cite{attention_is_all_you_need} is used to capture dependencies between any points within the input sequence, irrespective of their temporal distance, we used it for time series data, similar to~\cite{KDD_paper}.
Specifically, we used linear layers with a stack of four identical blocks.
The linear layer converts the input data to embedding vectors of 128. 
A token of size 128 is added to the embedded input as the representation vector. 
Each block is made up of a multi-head self-attention layer and a fully connected feed-forward layer with residual connections around them. 

\subsection{Self-Supervised Baseline Methods}
For our self-supervised learning experiments, we follow the same implementation setup as previous works~\cite{demirel2024unsupervisedapproachperiodicsource, KDD_paper, demirel2023chaos} for self-supervised learning in time series.
Specifically, we use a combination of convolutional with LSTM-based network, which is widely used in heart rate estimation from PPG signals~\cite{demirel2023chaos, CorNET}, as backbones for the encoder $f_{\theta}(.)$ where the projector is two fully connected layers. 
During pre-training, we use InfoNCE (for contrastive learning-based methods) as the loss function, which is optimized using Adam~\cite{Adam} with a learning rate of $0.003$.
We train the models with a batch size of 256 for 120 epochs and decay the learning rate using the cosine decay.
After pre-training, we train a single linear layer classifier on features extracted from the frozen pre-trained network, i.e., linear probing.

\paragraph{SimCLR}
SimCLR~\cite{SimCLR} introduces a contrastive learning framework for self-supervised visual representation learning. 
The method relies on maximizing agreement between differently augmented views of the same image via a contrastive loss in the latent space. 
We follow the previous implementations of SimCLR for time series~\cite{KDD_paper} and PPG signals~\cite{demirel2023chaos}.

\paragraph{NNCLR}
We follow a similar setup to SimCLR by applying two separate data augmentations, and then we use nearest neighbors in the learned representation space as the positive in contrastive losses~\cite{NNCLR}.
The maximum size of the support set equals 1024.

\paragraph{BYOL}
For the BYOL implementation, the exponential moving average parameter is set to 0.996 where the projector size is set to 128.
We set the learning rate to 0.03 similar to other SSL techniques.
Following the original implementation, we use a weight decay parameter of $1.5e-6$.

\paragraph{TS-TCC}
We follow the same architecture implementation with the losses, i.e., contextual and temporal contrasting.
TS-TCC~\cite{tstcc} proposed applying two separate augmentations, one augmentation is weak (jitter-and-scale) and the other is strong (permutation-and-jitter). 
The authors also change the strength of the permutation window from dataset to dataset. 
In our experiments, we follow the previous work~\cite{demirel2024unsupervisedapproachperiodicsource} for PPG-based augmentation.

\paragraph{TS2Vec} 
TS2Vec~\cite{TS2VEC} is an SSL method specifically designed for time series based on contrastive (instance and temporal-wise) learning in a hierarchical way over augmented context views where the context is generated by applying timestamp masking and random cropping on the input time series.
Following the original framework, we use a dilated CNN architecture with a depth of 10 and a hidden size of 64, which has a similar number of parameters to the architectures used by other SSL methods. 
The batch size is set to 256, and the number of epochs to 120, consistent with other SSL techniques.

\paragraph{VICReg}
We follow the original implementation and set the coefficients for each loss term to 25 ($\lambda$), 25 ($\mu$), and 1 ($\nu$), corresponding to the invariance, variance, and covariance terms, respectively.
We have not performed an additional hyper parameter search as these values are also set by previous on this application~\cite{demirel2024unsupervisedapproachperiodicsource}.

\paragraph{Barlow Twins}
Barlow Twins~\cite{barlow_twins} presents a function to avoid collapse for SSL by measuring the cross-correlation matrix between the outputs of two identical networks fed with augmented versions of a sample, and making it as close to the identity matrix as possible. 
This causes the embedding vectors of augmented versions of a sample to be similar while minimizing the redundancy between the components of these vectors.
Following the original implementation, we applied batch normalization to the extracted embeddings and set the hyperparameter $\lambda$ coefficient (in Equation~\ref{eq:barlow_twins}) to 0.005.

\begin{equation}\label{eq:barlow_twins}
\mathcal{L} = \sum_{i} (1 - C_{ii})^2 + \lambda \sum_{i} \sum_{j \neq i} C_{ij}^2,
\end{equation}

where \( C \) is the cross-correlation matrix computed between the two sets of normalized embeddings.

\subsection{Unsupervised Baseline Method}
We also consider one of the unsupervised learning based method, unsupervised periodicity detection (UPD), as an additional baseline model.

\paragraph{UPD} UPD~\cite{demirel2024unsupervisedapproachperiodicsource} presented two novel regularizers for detecting periodic patterns in time series data without using labels or specific augmentations.
We follow the original implementation \texttt{\href{https://github.com/eth-siplab/Unsupervised_Periodicity_Detection}{https://github.com/eth-siplab/Unsupervised\,\_\,Periodicity\,\_\,Detection}}.

Since this method operates without supervision, it requires relatively clean samples to learn periodic representations.
Consequently, its performance is quite low on datasets with noisy samples, such as WildPPG and DaLiA datasets.

\section{Additional Experiments}
Here, we present additional experiments we conducted beyond our main experiments, which focused on estimating the heart rate from green PPG recordings on the wrist as it is the most widely used measurement site and PPG wavelength for wearable devices~\cite{poor_fitbit}.

\subsection{Evaluation Across Multiple Body Locations and PPG Wavelengths}
We investigated the performance of the models when we integrate PPG signals from the different measurement sites of the body as well as PPG signals recorded at different wavelengths.
Table~\ref{appendix_tab:performance_ppg} presents the results for the baseline and the modified ResNet architecture performances.

\begin{table*}[h]
\centering
\caption{\label{appendix_tab:performance_ppg} Performance comparison of baselines with prior works in datasets}
\begin{minipage}{0.48\textwidth}
    \begin{adjustbox}{width=\textwidth,center}
    \renewcommand{\arraystretch}{1.35}
    \begin{tabular}{@{}llllllll@{}}
    \toprule
    \multirow{2}{*}{\begin{tabular}[c]{@{}c@{}}\textit{Location} \\ Method \end{tabular}} & \multicolumn{3}{l}{WildPPG} \\ 
    \cmidrule(r{15pt}){2-4} 
    & MAE$\downarrow$ & RMSE$\downarrow$ & $\rho$$\uparrow$ \\
    \hline
    \textit{Only Wrist} & & & \\
    1D ResNet & 8.62\small$\pm$0.06 & 15.00\small$\pm$0.06 & 42.41\small$\pm$0.24 \\
    Temp-ResNet    & 8.37\small$\pm$0.02 & 14.72\small$\pm$0.01 & 45.50\small$\pm$0.14 \\
    \hline
    \textit{Only Chest} & & & \\
    1D ResNet & 8.55\small$\pm$0.05 & 14.95\small$\pm$0.07 & 44.30\small$\pm$0.20 \\
    Temp-ResNet    & 8.35\small$\pm$0.03 & 14.68\small$\pm$0.02 & 46.48\small$\pm$0.10 \\
    \hline
    \textit{Wrist and Chest} & & & \\
    1D ResNet & 8.40\small$\pm$0.10 & 14.85\small$\pm$0.10 & 44.40\small$\pm$0.20 \\
    Temp-ResNet    & \textbf{8.28}\small$\pm$0.03 & \textbf{14.67}\small$\pm$0.02 & \textbf{47.55}\small$\pm$0.13 \\
    \hline
    \textit{All} & & & \\
    1D ResNet & 8.60\small$\pm$0.05 & 14.93\small$\pm$0.10 & 43.27\small$\pm$0.21 \\
    Temp-ResNet    & 8.33\small$\pm$0.05 & 14.70\small$\pm$0.05 & 46.50\small$\pm$0.10 \\
    \bottomrule
    \end{tabular}
    \end{adjustbox}
\end{minipage}
\hfill
\begin{minipage}{0.48\textwidth}
    \begin{adjustbox}{width=\textwidth,center}
    \renewcommand{\arraystretch}{1.3}
    \begin{tabular}{@{}llllllll@{}}
    \toprule
    \multirow{2}{*}{\begin{tabular}[c]{@{}c@{}}\textit{Wavelength} \\ Method \end{tabular}} & \multicolumn{3}{l}{WildPPG} \\ 
    \cmidrule(r{15pt}){2-4} 
    & MAE$\downarrow$ & RMSE$\downarrow$ & $\rho$$\uparrow$ \\
    \hline
    \textit{Green} & & & \\
    1D ResNet & 8.62\small$\pm$0.06 & 15.00\small$\pm$0.06 & 42.41\small$\pm$0.24 \\
    Temp-ResNet    & 8.37\small$\pm$0.02 & \textbf{14.72}\small$\pm$0.01 & 45.50\small$\pm$0.14 \\
    \hline
    \textit{Red} & & & \\
    1D ResNet & 10.25\small$\pm$0.14 & 17.63\small$\pm$0.21 & 35.21\small$\pm$0.23 \\
    Temp-ResNet    & 9.97\small$\pm$0.10 & 16.41\small$\pm$0.12 & 36.45\small$\pm$0.10 \\
    \hline
    \textit{Infrared} & & & \\
    1D ResNet & 8.97\small$\pm$0.12 & 15.93\small$\pm$0.12 & 36.33\small$\pm$0.16 \\
    Temp-ResNet    & 8.72\small$\pm$0.10 & 15.88\small$\pm$0.11 & 37.66\small$\pm$0.09 \\
    \hline
    \textit{All} & & & \\
    1D ResNet & 8.59\small$\pm$0.03 & 14.97\small$\pm$0.05 & 43.21\small$\pm$0.13 \\
    Temp-ResNet    & \textbf{8.35}\small$\pm$0.04 & 14.68\small$\pm$0.10 & \textbf{46.38}\small$\pm$0.10 \\
    \bottomrule
    \end{tabular}
    \end{adjustbox}
\end{minipage}
\end{table*}

From these results, we observe that incorporating additional measurement sites for HR prediction generally reduces the error, despite increasing the number of model parameters.
Furthermore, results suggest that the wrist measurement site has the lowest signal quality compared to other sites.
This outcome is expected, as the wrist is particularly susceptible to motion artifacts and exposure to cold, and the results are consistent with prior work~\cite{longmore_comparison_2019}.
Similarly, the signal quality of alternative wavelengths is lesser compared to green PPG which is also consistent with prior work~\cite{ray_review_2023}.

\subsection{Cross-Dataset Evaluation}

We evaluated model performance using a cross-dataset approach, specifically employing a leave-one-dataset-out scheme. 
In this setup, two datasets were used for training, while the third dataset was reserved for testing. 
Results are summarized in Table~\ref{tab:performance_ppg_cross_dataset}.

\begin{table*}[h]
\centering
\caption{Performance comparison of baselines in cross-dataset evaluation (one dataset left for evaluation and others used for training)}
\begin{adjustbox}{width=1\columnwidth,center}
\renewcommand{\arraystretch}{1.1}
\begin{tabular}{@{}lllllllllll@{}}
\toprule
\multirow{2}{*}{Method} & \multicolumn{3}{l}{Test on WildPPG} & \multicolumn{3}{l}{Test on SPC12\tnote{*}} & \multicolumn{3}{l}{Test on DaLiA} \\ 
\cmidrule(r{15pt}){2-4}  \cmidrule(r{15pt}){5-7}  \cmidrule(r{15pt}){8-10}
& MAE$\downarrow$ & RMSE$\downarrow$ & $\rho$$\uparrow$ & MAE$\downarrow$ & RMSE$\downarrow$ & $\rho$$\uparrow$ & MAE$\downarrow$ & RMSE$\downarrow$ & $\rho$$\uparrow$ \\
\midrule
\textit{Supervision} & & & & & & & & & \\
1D ResNet & \underline{16.32}\small$\pm$0.26 & \underline{27.33}\small$\pm$0.33 & \underline{12.33}\small$\pm$0.24 & \textbf{11.25}\small$\pm$1.78 & \textbf{18.37}\small$\pm$0.88 & \textbf{67.36}\small$\pm$0.78 & \underline{17.23}\small$\pm$2.28 & \underline{22.10}\small$\pm$0.13 & \underline{67.14}\small$\pm$0.15\\
DCL      & 19.43\small$\pm$2.12 & 29.32\small$\pm$2.43 & 10.68\small$\pm$2.59 & 15.45\small$\pm$2.34 & 26.32\small$\pm$2.09 & 55.43\small$\pm$4.75 & 20.90\small$\pm$0.55 & 25.31\small$\pm$1.54 & 60.21\small$\pm$2.56 \\
FCN      & 16.55\small$\pm$1.23 & 26.77\small$\pm$0.26 & 11.53\small$\pm$1.56 & 14.22\small$\pm$2.45 & 24.58\small$\pm$1.96 & 50.28\small$\pm$1.50 & 24.13\small$\pm$1.13 & 24.56\small$\pm$2.31 & 65.32\small$\pm$2.56 \\
LSTM & 16.79\small$\pm$1.30 & 28.57\small$\pm$2.43 & 8.97\small$\pm$1.10 & 27.86\small$\pm$2.21 & 28.11\small$\pm$4.10 & 42.11\small$\pm$3.95 & 21.59\small$\pm$1.10 & 23.47\small$\pm$1.67 & 63.42\small$\pm$1.30 \\
Transformer   & 17.25\small$\pm$1.45 & 30.98\small$\pm$2.59 & 6.72\small$\pm$3.66 & 28.45\small$\pm$1.98 & 30.38\small$\pm$3.59 & 30.85\small$\pm$2.05 & 20.31\small$\pm$2.56 & 20.17\small$\pm$2.17 & 55.21\small$\pm$2.17 \\
\bottomrule
\end{tabular}
\end{adjustbox}
\label{tab:performance_ppg_cross_dataset}
\end{table*}

\subsection{Additional sensor modality}
In Table~\ref{tab:performance_ppg_imu} we present the experimental results of integrating acceleration with PPG as an additional modality.

\begin{table*}[h]
\centering
\caption{Performance comparison of baseline models across datasets when integrating accelerometer data with PPG as an additional modality}
\begin{adjustbox}{width=1\columnwidth,center}
\renewcommand{\arraystretch}{1.1}
\begin{tabular}{@{}lllllllllll@{}}
\toprule
\multirow{2}{*}{Method} & \multicolumn{3}{l}{Test on WildPPG} & \multicolumn{3}{l}{Test on SPC12\tnote{*}} & \multicolumn{3}{l}{Test on DaLiA} \\ 
\cmidrule(r{15pt}){2-4}  \cmidrule(r{15pt}){5-7}  \cmidrule(r{15pt}){8-10}
& MAE$\downarrow$ & RMSE$\downarrow$ & $\rho$$\uparrow$ & MAE$\downarrow$ & RMSE$\downarrow$ & $\rho$$\uparrow$ & MAE$\downarrow$ & RMSE$\downarrow$ & $\rho$$\uparrow$ \\
\midrule
\textit{Supervision} & & & & & & & & & \\
1D ResNet & \underline{8.22}\small$\pm$0.09 & \underline{14.34}\small$\pm$0.07 & \underline{44.33}\small$\pm$0.22 & \textbf{5.33}\small$\pm$0.18 & \textbf{11.08}\small$\pm$0.67 & \textbf{85.02}\small$\pm$0.47 & \underline{4.43}\small$\pm$0.10 & \underline{09.88}\small$\pm$0.13 & \underline{85.96}\small$\pm$0.20\\
DCL      & 8.69\small$\pm$0.03 & 15.03\small$\pm$0.09 & 42.01\small$\pm$0.07 & 15.30\small$\pm$1.20 & 20.98\small$\pm$1.33 & 20.78\small$\pm$1.26 & 4.88\small$\pm$0.12 & 11.73\small$\pm$0.07 & 79.12\small$\pm$0.12 \\
FCN      & 9.65\small$\pm$0.20 & 15.16\small$\pm$0.22 & 37.42\small$\pm$0.71 & 25.95\small$\pm$0.37 & 27.87\small$\pm$0.12 & 45.13\small$\pm$1.80 & 6.12\small$\pm$0.23 & 10.92\small$\pm$0.28 & 84.04\small$\pm$0.03 \\
LSTM & 8.72\small$\pm$0.20 & 14.11\small$\pm$0.10 & 38.61\small$\pm$1.10 & 17.41\small$\pm$2.30 & 21.13\small$\pm$2.57 & 60.10\small$\pm$3.12 & 5.07\small$\pm$0.08 & 10.94\small$\pm$0.20 & 80.01\small$\pm$0.30 \\
Transformer   & 9.98\small$\pm$0.15 & 17.10\small$\pm$0.23 & 31.73\small$\pm$1.08 & 21.92\small$\pm$0.20 & 25.10\small$\pm$0.18 & 50.97\small$\pm$0.21 & 7.17\small$\pm$0.15 & 14.98\small$\pm$0.20 & 68.97\small$\pm$0.66 \\
Temp-ResNet    & \textbf{8.13}\small$\pm$0.03 & \textbf{13.95}\small$\pm$0.02 & \textbf{47.48}\small$\pm$0.10 & --- & ---&---& \textbf{4.27}\small$\pm$0.03 & \textbf{9.13}\small$\pm$0.15 & \textbf{87.31}\small$\pm$0.19 \\
\bottomrule
\end{tabular}
\end{adjustbox}
\label{tab:performance_ppg_imu}
\end{table*}

\subsection{Baseline Results on the BIDMC Dataset}

Considering the wide use of the BIDMC dataset (recorded in controlled environments), we provide baseline results of the baseline methods listed in Section~\ref{appendix_baselines} for this dataset in Table~\ref{tab:performance_bidmc}.
\begin{table*}[h]
\centering
\caption{Performance comparison of baselines when tested on the BIDMC dataset}
\begin{adjustbox}{width=0.5\columnwidth,center}
\renewcommand{\arraystretch}{1.1}
\begin{tabular}{@{}lllll@{}}
\toprule
\multirow{2}{*}{Method} & \multicolumn{3}{l}{BIDMC} \\ 
\cmidrule(r{15pt}){2-4} 
& MAE$\downarrow$ & RMSE$\downarrow$ & $\rho$$\uparrow$ \\
\midrule
\textit{Heuristic} & & & \\
FFT & 4.41 & 9.74 & 33.42 \\
\midrule
\textit{Supervision} & & & \\
1D ResNet & \textbf{3.62}\small$\pm$0.22 & \textbf{5.23}\small$\pm$0.39 & \underline{85.43}\small$\pm$0.20 \\
DCL      & 4.18\small$\pm$0.31 & 5.89\small$\pm$0.65 & 83.34\small$\pm$1.37 \\
FCN      & 4.01\small$\pm$0.27 & 5.41\small$\pm$0.47 & 84.57\small$\pm$1.14 \\
LSTM     & 5.73\small$\pm$1.79 & 10.86\small$\pm$0.96 & 29.37\small$\pm$1.58 \\
Transformer & 7.71\small$\pm$1.40 & 11.03\small$\pm$1.35 & 19.21\small$\pm$2.20 \\
\bottomrule
\end{tabular}
\end{adjustbox}
\label{tab:performance_bidmc}
\end{table*}

\section{Impact of Temperature and Motion}
\begin{figure}[!h]
    \centering
    \includegraphics[width=\columnwidth]{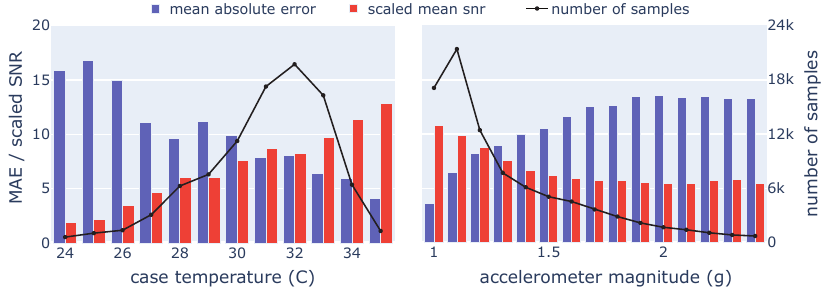}
    \caption{The error (blue) of 1D ResNet based on PPG from the wrist, dependent on the device temperature (left), and device motion (right). Inversely related is the computed SNR of the PPG signal (red).}
    \label{fig:temp_imu_error}
\end{figure}

The modified architecture Temp-ResNet adds the device's case temperature (measured within the encasing) as an additional input modality to improve results.
Figure~\ref{fig:temp_imu_error} shows the HR error of 1D ResNet based on PPG from the wrist in relation to case temperature and motion.
The correlations of error and temperature/motion are supported by the investigation of signal-to-noise ratio (SNR):
We have computed the SNR of the PPG signal by relating the power $P_{HR}$ of the ground truth HR in the frequency spectrum of the PPG signal to the power $P_{noise} = P_{tot}-P_{HR}$ of the other frequency components in the relevant window of possible heart rates:

\[SNR = 10\cdot log \left( \frac{P_{HR}}{P_{tot}-P_{HR}} \right) dB\]

While it is well understood that motion artifacts have a negative impact on PPG signal quality~\cite{TROIKA}, the presented results show that the temperature within the wearable has an inverse correlation to PPG signal quality as well.

\section{PPG Device}
The schematics and production data (CAD drawing, BOM, Gerber/NCDrill, pick\&place) of the PPG device used in this project are available on the project website. The data is acquired using an off-the-shelf PPG LED module (SFH7072, OSRAM) and a commonly used dedicated PPG acquisition chip (MAX86141, Analog Devices). Table~\ref{tab:max86141} shows the register configuration of the MAX86141 PPG chip as used during the data acquisition.

\begin{table*}[h]
\caption{Register configurations of the MAX86141 PPG chip.}
\begin{adjustbox}{width=1\columnwidth,center}
\begin{tabular}{llll}
Register Name                  & Reg Addr & Value & Description                                        \\ \hline
MAX86141\_PPG\_CONFIG\_1       & 0x11     & 0x2B  & ALC enable, 117us integration time, 16uA ADC range \\
MAX86141\_PPG\_CONFIG\_2       & 0x12     & 0x70  & 128 Hz SR (ext. clock), no averaging               \\
MAX86141\_PPG\_CONFIG\_3       & 0x13     & 0xC0  & 12uS LED settle time                               \\
MAX86141\_PHOTO\_DIODE\_BIAS   & 0x15     & 0x11  & Low capacity PD                                    \\
MAX86141\_LED\_RANGE\_1        & 0x2A     & 0x00  & LED driver range 31mA                              \\
MAX86141\_LED1\_PA             & 0x23     & 0x12  & IR LED current 2.16mA                              \\
MAX86141\_LED2\_PA             & 0x24     & 0x0C  & R LED current 1.44mA                               \\
MAX86141\_LED3\_PA             & 0x25     & 0x0C  & G LED current 1.44mA                               \\
MAX86141\_INTERRUPT\_ENABLE\_1 & 0x02     & 0x00  & Disable all Interrupts                             \\
MAX86141\_LED\_SEQ\_REG1       & 0x20     & 0x21  & Set LED sequence (1 slot each for IR R G)          \\
MAX86141\_LED\_SEQ\_REG2       & 0x21     & 0x03  & Set LED sequence (1 slot each for IR R G)          \\
MAX86141\_LED\_SEQ\_REG3       & 0x22     & 0x00  & Set LED sequence (1 slot each for IR R G)         
\end{tabular}
\label{tab:max86141}
\end{adjustbox}
\end{table*}

\section{File Structure}
\begin{figure}[!h]
    \centering
    \includegraphics[width=\columnwidth]{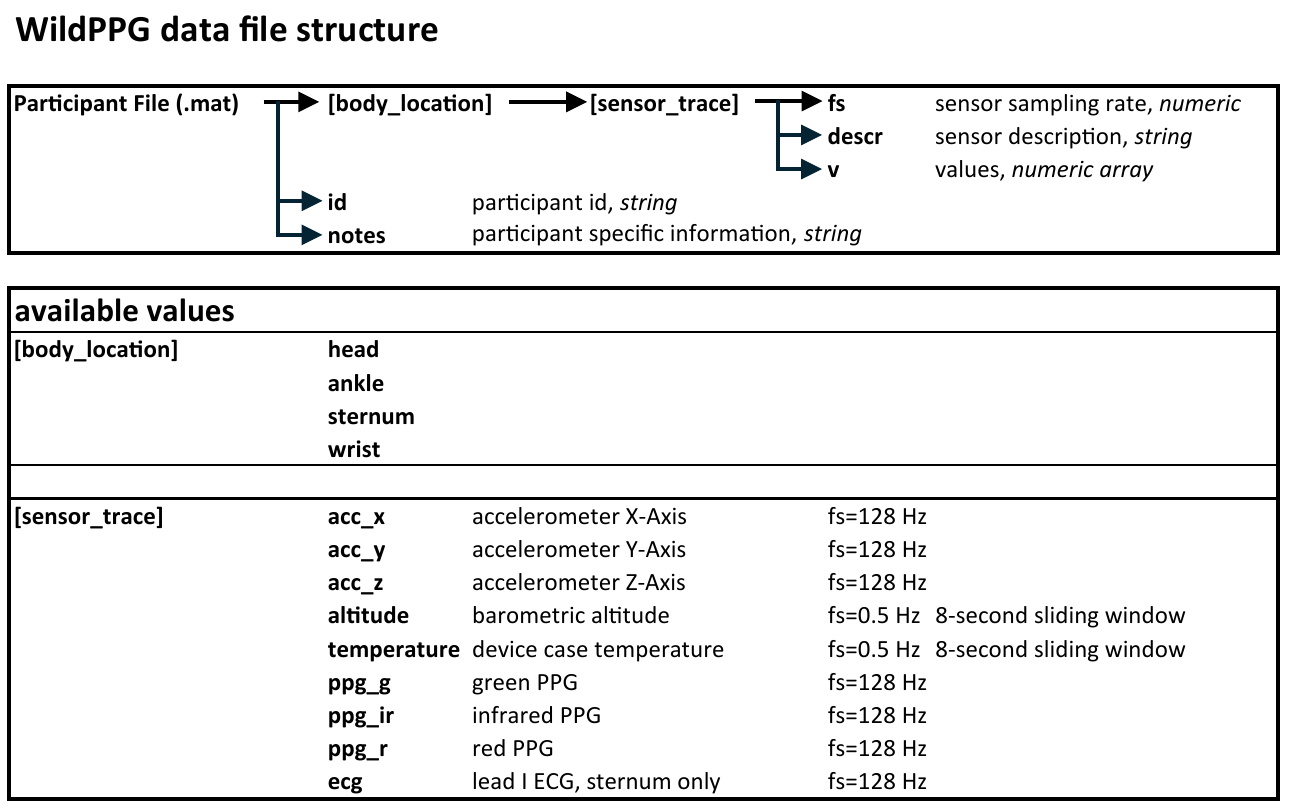}
    \caption{For each participant in \projname, one .mat Matlab data file is provided with the structure as illustrated.}
    \label{fig:filestructure}
\end{figure}

The file structure of the \projname dataset is shown in Figure~\ref{fig:filestructure}

\section{Author Statement}
The authors bear all responsibility in case of violation of rights.

\end{document}